\documentclass[journal]{IEEEtran}

\usepackage{times}
\usepackage{mathrsfs}
\usepackage{epsfig}
\usepackage{graphicx}
\usepackage{amsmath}
\usepackage{amssymb}
\usepackage{graphicx}
\usepackage{cite}
\usepackage{enumerate}
\usepackage{cases}
\usepackage{multirow}
\usepackage{verbatim}
\usepackage{amssymb}
\usepackage{CJK}
\usepackage{algorithm}
\usepackage{algorithmicx}
\usepackage{algpseudocode}
\usepackage{color}
\usepackage{bm}
\usepackage{booktabs}
\usepackage{amssymb}

\usepackage[pagebackref=true,breaklinks=true,letterpaper=true,colorlinks,bookmarks=false]{hyperref}

\usepackage{color}
\usepackage{tikz}
\usepackage{ifthen}
\usepackage{bm}
\usetikzlibrary{3d,decorations.text,shapes.arrows,positioning,fit,backgrounds, intersections}
\usetikzlibrary{shapes.misc}
\definecolor{rdcolor}{RGB}{171, 197, 196}
\definecolor{ddcolor}{RGB}{173, 185, 202}

\definecolor{color3}{RGB}{249, 205, 173}
\definecolor{color4}{RGB}{229, 131, 8}
\definecolor{color7}{HTML}{BF7130}
\definecolor{color9}{HTML}{A5B92E}

\definecolor{goldenrod}{RGB}{85, 65, 13}
\definecolor{navyblue}{RGB}{0, 0, 50}
\definecolor{dandelion}{RGB}{94, 88, 19}
\definecolor{brickred}{RGB}{80, 25, 33}
\definecolor{rfcolor}{HTML}{FF4500}
\definecolor{dfcolor}{HTML}{61D8A2}
\definecolor{gmacolor}{RGB}{237, 147, 147}
\definecolor{gmaedge}{RGB}{227, 91, 91}
\definecolor{fcolor}{RGB}{251, 241, 117}
\definecolor{fedge}{RGB}{217, 212, 0}
\definecolor{mfcolor}{RGB}{255, 116, 0}
\definecolor{mfEdge}{RGB}{237, 107, 55}
\definecolor{linecolor}{RGB}{16, 73, 94}

\tikzset{drawLine/.style={->,  line width=1.5pt, color=linecolor!55}}
\tikzset{rfLine/.style={->, line width=1.5pt,color=rfcolor!50}}
\tikzset{dfLine/.style={->, line width=1.5pt,color=dfcolor!50}}
\definecolor{edgeColor}{RGB}{32, 32, 32}
\tikzset{circle dotted/.style={dash pattern=on .05mm off 2mm,
		line cap=round}}
\tikzset{global scale/.style={
		scale=#1,
		every node/.append style={scale=#1}
	}}

\tikzstyle{GMASTYLE}=[draw, rectangle, color=gmaedge, text=black, minimum height=6mm, minimum width=7mm/0.618, fill=gmacolor!60, rounded corners=4pt, line width=.5pt]
\tikzstyle{FSTYLE}=[draw, circle,color=fedge, minimum width=4mm, inner sep=0.1mm, fill=fcolor!60,  text=black, scale=0.9]
\tikzstyle{SSTYLE}=[draw, circle, minimum size=12pt, inner sep=0pt, fill=navyblue!30]
\tikzstyle{ASTYLE}=[draw, circle, minimum size=12pt, inner sep=0pt, fill=pink!30]
\tikzstyle{MFSTYLE}=[draw, rounded rectangle, color=mfEdge, minimum height=6mm,fill=mfcolor!45, text=black]
\usepackage{tkz-euclide}
\tikzset{
	connect/.style args={(#1) to (#2) over (#3) by #4}{
		insert path={
			let \p1=($(#1)-(#3)$), \n1={veclen(\x1,\y1)},
			\n2={atan2(\y1,\x1)}, \n3={abs(#4)}, \n4={#4>0 ?180:-180}  in
			(#1) -- ($(#1)!\n1-\n3!(#3)$)
			arc (\n2:\n2+\n4:\n3) -- (#2)
		}
	},
}
\tikzset{dist/.style={path picture= {
			\begin{scope}[x=1pt,y=10pt]
				\draw plot[domain=-6:6] (\x,{1/(1 + exp(-\x))-0.5});
			\end{scope}
		}
	}
}
				\newcommand{\networkLayer}[8]{
					\def\a{#1} 
					\def\b{0.02}
					\def\c{#2} 
					\def\t{#3} 
					\def\d{#4} 
					\def\z{#5} 
					\def\round{0.4pt}
					\def\linept{2pt}
					
					\ifthenelse{\equal{#8}{}}{
						\draw[line width=\linept, rounded corners=\round, color=edgeColor](\c+\t,\z,\d) -- (\c+\t,\a+\z,\d) --(\t,\a+\z,\d); 	
					}{
					\draw[line width=\linept, rounded corners=\round, color=edgeColor](\c+\t,\z,\d) -- (\c+\t,\a+\z,\d) --(\t,\a+\z,\d)node[midway, above]{#8}; 				
				} 
				\draw[line width=\linept, rounded corners=\round, color=edgeColor](\t,\z,\a+\d) -- (\c+\t,\z,\a+\d) node[midway,below] {#7} -- (\c+\t,\a+\z,\a+\d) -- (\t,\a+\z,\a+\d)  -- (\t,\z,\a+\d) --cycle; 
				\draw[line width=\linept, rounded corners=\round, color=edgeColor](\c+\t,\z,\d) -- (\c+\t,\z,\a+\d) --(\c+\t, \z+\a, \a+\d) -- (\c+\t, \a+\z, \d)--cycle;
				\draw[line width=\linept, rounded corners=\round, color=edgeColor](\c+\t,\a+\z,\d) -- (\c+\t,\a+\z,\a+\d);
				\draw[line width=\linept, rounded corners=\round, color=edgeColor](\t, \a+\z, \d) -- (\t, \a+\z, \a+\d)-- (\c+\t, \a+\z, \a+\d)-- (\t+\c, \a+\z, \d)--cycle; 

				\filldraw[#6,line width=0pt,rounded corners=\round] (\t+\b,\b+\z,\a+\d) -- (\c+\t-\b,\b+\z,\a+\d) -- (\c+\t-\b,\a-\b+\z,\a+\d) -- (\t+\b,\a-\b+\z,\a+\d) ; 
				
				\filldraw[#6,line width=0pt,rounded corners=\round] (\t+\b,\a+\z,\a-\b*4+\d) -- (\c+\t-\b*2,\a+\z, \a-\b*4+\d) -- (\c+\t-\b*2,\a+\z,\b+\d) -- (\t+\b,\a+\z,\b+\d); 
				\ifthenelse {\equal{#6} {}}
				{} 
				{\filldraw[#6,line width=0pt, rounded corners=\round] (\c+\t,\b*2+\z,\a-\b*4+\d-\b) -- (\c+\t,\b*2+\z,\b+\d) -- (\c+\t,\a-\b*2.68+\z,\b+\d) -- (\c+\t,\a-\b*2.68+\z,\a-\b*4+\d-\b);} 
			}
\usepackage{multirow}
\def\metrics{$F_{\beta}\uparrow$ \quad {  }$S_{m} \uparrow$ \quad MAE $\downarrow$ }
\def\triplets(#1,#2,#3){#1 \quad #2\quad#3}
\def\tripletsr(#1,#2,#3){\textcolor{red}{#1}\quad  \textcolor{red}{#2}\quad       \textcolor{red}{#3}}
\def\fillcell{$\text{--}\phantom{0.8888}\text{--}\phantom{0.8888}\text{--}$}
\usepackage{enumitem}

\usepackage{booktabs}

\newcommand{\etal}{\textit{et al}.}
\newcommand{\ie}{\textit{i}.\textit{e}.}
\newcommand{\eg}{\textit{e}.\textit{g}.}

\begin{document}

\title{PSNet: Parallel Symmetric Network for Video Salient Object Detection}

\author{Runmin Cong,~\IEEEmembership{Member,~IEEE,} Weiyu Song, Jianjun Lei,~\IEEEmembership{Senior Member,~IEEE,} Guanghui Yue,\\ Yao Zhao,~\IEEEmembership{Senior Member,~IEEE,} and Sam Kwong,~\IEEEmembership{Fellow,~IEEE}

\thanks{Runmin Cong is with the Institute of Information Science, Beijing Jiaotong University, Beijing 100044, China, also with the Beijing Key Laboratory of Advanced Information Science and Network Technology, Beijing 100044, China, and also with the Department of Computer Science, City University of Hong Kong, Hong Kong SAR, China (e-mail: rmcong@bjtu.edu.cn).}
\thanks{Weiyu Song and Yao Zhao are with the Institute of Information Science, Beijing Jiaotong University, Beijing 100044, China, and also with the Beijing Key Laboratory of Advanced Information Science and Network Technology, Beijing 100044, China (e-mail: wysong125@bjtu.edu.cn, yzhao@bjtu.edu.cn).}
\thanks{Jianjun Lei is with the School of Electrical and Information Engineering, Tianjin University, Tianjin 300072, China
(e-mail: jjlei@tju.edu.cn).}
\thanks{Guanghui Yue is with the National, regional Key Technology Engineering Laboratory for Medical Ultrasound, Guangdong Key Laboratory for Biomedical Measurements and Ultrasound Imaging, School of Biomedical Engineering, Health Science Center, Shenzhen University, Shenzhen 518060, China (email: yueguanghui@szu.edu.cn).}
\thanks{Sam Kwong is with the Department of Computer Science, City University of Hong Kong, Hong Kong SAR, China, and also with the City University of Hong Kong Shenzhen Research Institute, Shenzhen 51800, China (e-mail: cssamk@cityu.edu.hk).}
}

\markboth{}
{Shell \MakeLowercase{\textit{\etal}}: Bare Demo of IEEEtran.cls for IEEE Journals}
\maketitle

\begin{abstract}
For the video salient object detection (VSOD) task, how to excavate the information from the appearance modality and the motion modality has always been a topic of great concern. The two-stream structure, including an RGB appearance stream and an optical flow motion stream, has been widely used as a typical pipeline for VSOD tasks, but the existing methods usually only use motion features to unidirectionally guide appearance features or adaptively but blindly fuse two modality features. However, these methods underperform in diverse scenarios due to the uncomprehensive and unspecific learning schemes. In this paper, following a more secure modeling philosophy, we deeply investigate the importance of appearance modality and motion modality in a more comprehensive way and propose a VSOD network with up and down parallel symmetry, named PSNet. Two parallel branches with different dominant modalities are set to achieve complete video saliency decoding with the cooperation of the Gather Diffusion Reinforcement (GDR) module and Cross-modality Refinement and Complement (CRC) module. Finally, we use the Importance Perception Fusion (IPF) module to fuse the features from two parallel branches according to their different importance in different scenarios. Experiments on four dataset benchmarks demonstrate that our method achieves desirable and competitive performance. The code and results can be found from the link of \url{https://rmcong.github.io/proj\_PSNet.html}.
\end{abstract}


\section{Introduction} \label{sec1}
\IEEEPARstart{V}{ideo} salient object detection (VSOD) focuses on extracting the most attractive and motion related objects in a video sequence \cite{DBLP:journals/pami/WangLFSLY22,crm/tcsvt19/review}, which has been used as a pre-processing step for a wide range of tasks, such as video understanding \cite{intro-3,DBLP:journals/pami/WangSLHL21,crm/ACMMM20/DMVOS,wang2016stereoscopic}, video compression \cite{itti2004automatic}, video tracking \cite{DBLP:conf/iccv/ZhouP0WZ021}, and video caption\cite{li2019visual}.   
Due to the characteristic of video,  in addition to the appearance cue, the motion attribute plays an important role, which is different from the SOD task for static images. 
Entering the deep learning era,  a variety of VSOD methods have been explored, which can be roughly divided into two categories, \eg, single-stream methods using the temporal convolution/long short-term
memory  \cite{wang2017video,gu2020pyramid, chen2021exploring,song2018pyramid, fan2019shifting} and two-stream methods using the optical flow \cite{su2020ds,li2019motion,ji2021full,chen2021confidence}. Even so, it is still very challenging for current VSOD methods to fully excavate and integrate the information from motion and appearance cues. 
\IEEEpeerreviewmaketitle
 \begin{figure}[t]
 	\centering
 	\includegraphics[width=\columnwidth]{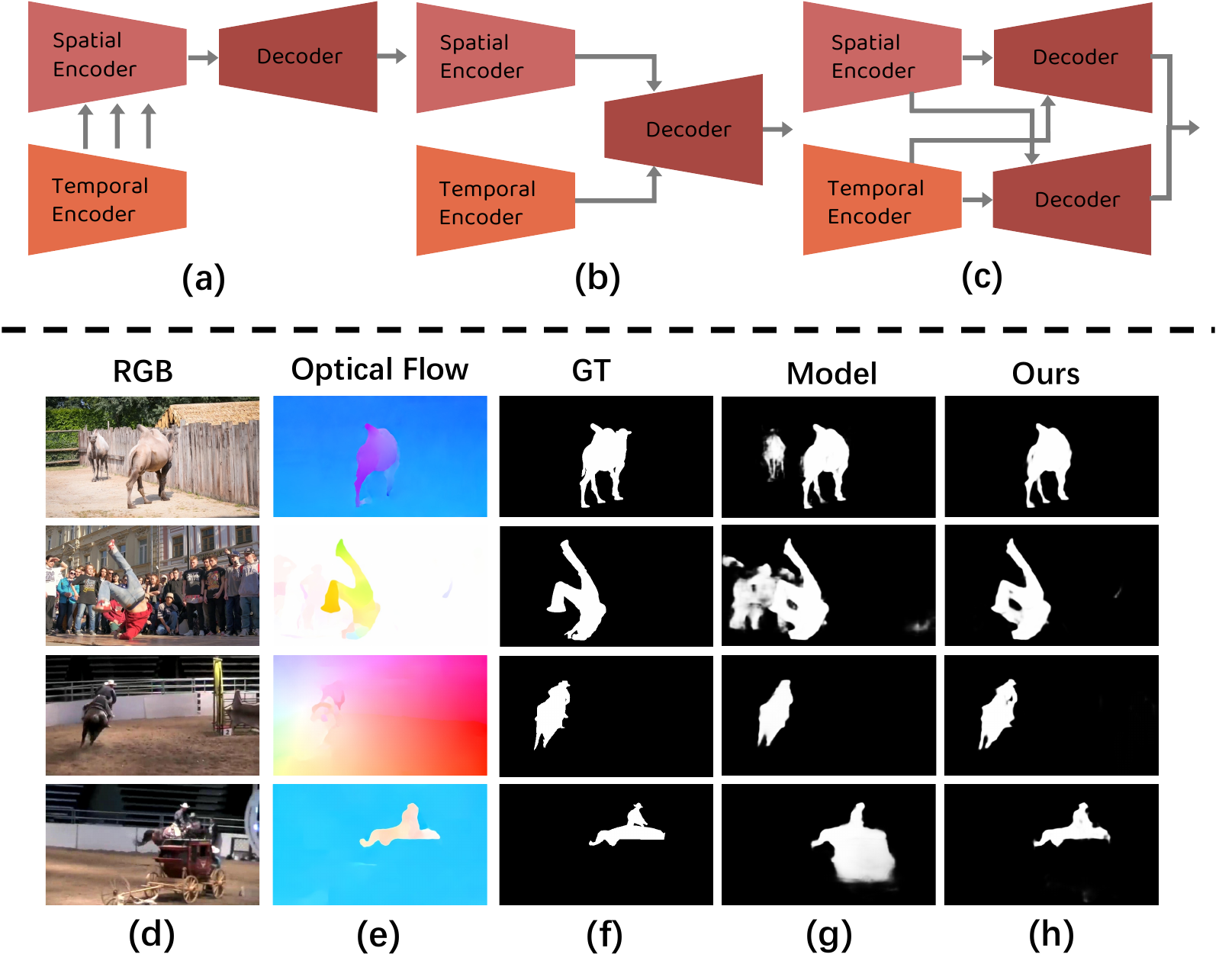}
 	\caption {Top: The structures of VSOD models between our method (c) and the other optical flow-based two-stream VSOD methods (a) (b). Bottom: The saliency results from different models in different scenes. (d) RGB images; (e) Optical flow images; (f) GT; (g) Saliency maps deduced by different methods, where the first row is generated by the MGA method\cite{li2019motion}, the second row is generated by our baseline model with addition fusion, and the last two rows are generated by the CAG method \cite{chen2021confidence}; (h) Our model.
 	}
 	\label{fig1}
 \end{figure}
For the optical flow-based two-stream VSOD model, how to achieve the information interaction according to the role of the two modalities is very important. 
In this paper, we first rethink and review the interaction mode in the optical flow-based two-stream VSOD structure, and the existing methods can be further categorized into two categories. One is the unidirectional guidance model, as shown in Fig. \ref{fig1}(a), in which the motion information mainly plays a supplementary role.
For example, Li \etal \cite{li2019motion} encouraged motion features to guide the appearance features in the designed VSOD model. As a result, the model pays too much attention to the spatial branch, while the advantage of the motion branch is weakened when dealing with some challenging scenes, such as the stationary objects with salient appearance may be incorrectly preserved. (see the $1^{st}$ row of Fig. \ref{fig1}). 
To alleviate the problems mentioned above, the undifferentiated and bidirectional fusion mechanism is proposed as another typical interaction mode,  as shown in Fig. \ref{fig1}(b), which no longer distinguishes their primary and secondary roles. 
Fusing the two modality features by addition or concatenation is the simplest solution, but this way often fails to achieve the desired results, especially for some complex scenes (see the $2^{nd}$ row of Fig. \ref{fig1}). In addition, some works \cite{chen2021confidence} learn the weights to determine the contributions of spatial and temporal features, and then achieve adaptive fusion of two modality features. 
Although these methods appear to be quite intelligent and achieve relatively competitive performance, this black-box adaptive fusion strategy sometimes only trades off performance rather than maximizing gains when faced with different scenarios.
As shown in $3^{rd}$ and $4^{th}$ rows of Fig. \ref{fig1}, they are different frames from different moments of the same video. Although they are similar scenes, the contribution of the two modality data to the final saliency detection is different. We can find that the appearance cues are more important than the motion cues in the $3^{rd}$ row, where the dramatic moving of objects and the change of camera position lead to unclear and blur motion cues. While in the $4^{th}$ row, motion cues can provide more effective guidance information compared with appearance cues that contain some irrepressible noise. 
According to these observations, when salient objects and backgrounds share similar appearances or background interference is disturbing, interlaced and wrong appearance cues could greatly contaminate the final detection results. But at this time, perhaps accurate motion cues will help us to segment the salient objects correctly. Alternatively, too slow or too fast object motion will blur the estimated optical flow map, thus failing to provide discriminative motion cues and affecting the final detection. In this case, satisfactory detection results can be obtained by exploiting the semantic information from distinctive appearance cues and features. In other words, the roles of the two modalities in different scenes or even similar scenes cannot be generalized, and the uncertainty of the scene makes it very difficult to model interaction fully adaptively.
Instead of learning the importance of these two modalities regardless and fully adaptively, we propose a more secure modeling strategy, where the importance of appearance cues and motion cues will be comprehensively and explicitly taken into account to generate the saliency maps, as shown in Fig. \ref{fig1}(c). In our network, we design a top-bottom parallel symmetric structure, which sacrifices the full-automatic intelligence so that we can fuse features more comprehensively, considering the adaptability of the network to different scenarios. Since it struggles for the network to distinguish which modality is more important in one particular scenario, we design two branches with varying tendencies of importance for VSOD, taking one modality feature as a dominant role in each branch and then supplementing from another modality. 


\begin{figure*}[t]
	 \centering
	\includegraphics[width=0.98\textwidth]{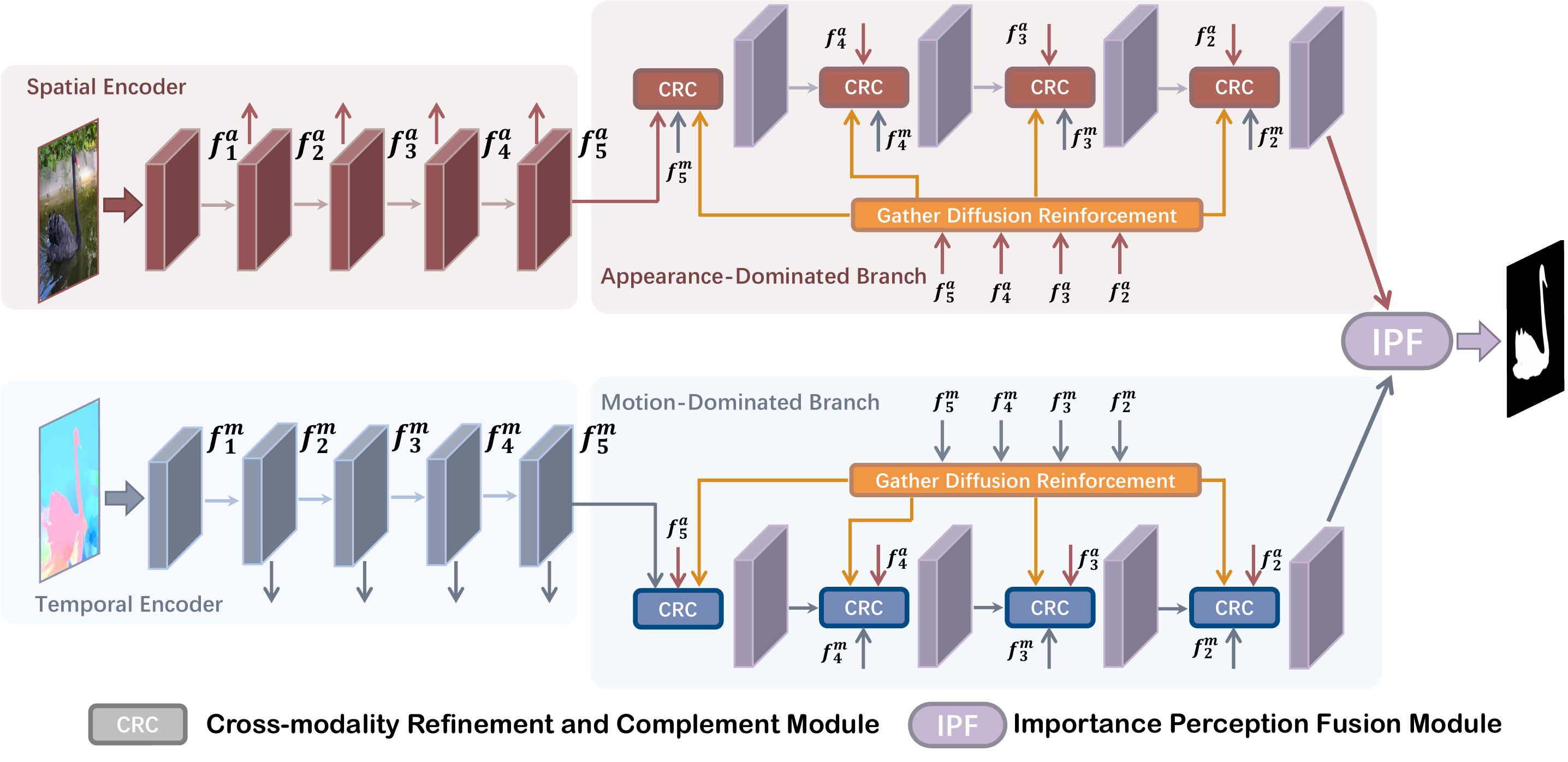}	
	
	\caption{The flowchart of the proposed Parallel Symmetric Network (PSNet) for video salient object detection. We first extract the multi-level features from RGB images and optical flow maps via spatial encoder and temporal encoder  respectively, which are denoted as $f_i^{a}$ and $f_i^{m}$ ($i=\{1,2,\cdots,5\}$). Then, the appearance-dominated branch (top branch) and motion-dominated branch (bottom branch) are used to feature decoding.
	For each decoding, we use Gather Diffusion Reinforcement (GDR) module to perform cross-scale feature enhancement, and then use the Cross-modality Refinement and Complement (CRC) module to achieve cross-modality interaction with an explicit primary and secondary modality relationship.
	Finally, the Importance Perception Fusion (IPF) module is used to integrate the upper and lower branches by considering their different importance in different scenarios. }
	\label{fig2}
\end{figure*}

\indent Under the parallel symmetric structure, we need to do two things, one is how to realize the utilization of the two modality information in each branch more clearly, and the other is how to integrate the information of the upper and lower branches to generate the final result. For the first issue, we design the Gather Diffusion Reinforcement (GDR) module and Cross-modality Refinement and Complement (CRC) module to achieve dominate-modality feature reinforcement and cross-modality feature interaction, respectively. Considering that the high-level semantic information can reduce the interference of non-salient information in a single modality and multi-scale information can contribute to more comprehensive features, we design a GDR module to enhance the effectiveness of dominant features in each branch and improve the multi-scale correlation of the dominant features themselves. The outputs of the GDR module are then used for the CRC module in a top-down manner. The key ideas behind the design of the CRC module are as follows. Even if the data from one modality plays a dominant role, there is more or less useful information from the other modality. We divide this role into two types, one is the refinement role, which is mainly used to suppress the irrelevant redundancies in the dominant features, and the other is the complementary role, mainly used to compensate for potential information missing in dominant features. Therefore, we design the CRC module to achieve comprehensive information interaction in the case of explicit primary and secondary relations, which can play the most significant role in our proposed parallel symmetric framework. Although both our upper and lower branches are fully implemented in the VSOD task, the dominant modality they set is different. To obtain more robust and generalized final results, we need to integrate the two branches, which is the second problem we need to solve. Considering the different importance of the upper and lower branches in different scenarios, we 
introduce an Importance Perception Fusion (IPF) module for adaptive fusion. All designed modules are closely cooperated and integrated under our parallel symmetrical structure to achieve better detection performance. As shown in the $5^{th}$ column of Fig. \ref{fig1}, our model can accurately locate salient objects in different types of scenes, with obvious advantages in detail representation and background suppression.
The contributions of this paper can be summarized as:
\begin{itemize}[noitemsep, topsep=0pt]
	\item Considering the adaptability of the network to different scenarios and the uncertainty of the role of different modalities, we propose a parallel symmetric network (PSNet) for VSOD that simultaneously models the importance of two modality features in an explicit way. 
	\item We propose a GDR module in each branch to perform multi-scale content enhancement for dominant features and design a CRC module to achieve cross-modality interaction, where the auxiliary features are applied to refine and supplement dominant features.
	\item Experimental results on four mainstream datasets demonstrate that our PSNet outperforms 25 state-of-the-art methods both quantitatively and qualitatively.
\end{itemize}

\section{Related Work}
\subsection{Salient Object Detection in Single Image and Image Group}
For decades, single image-based SOD task has achieved extensive development \cite{DBLP:journals/tip/WangS18,DBLP:journals/pami/WangSDBY20,crm/aaai20/GCPANet,liu2018picanet,zhao2019egnet, qin2019basnet, wang2017edge,pang2020multi,fang2022densely,crm/tip21/DAFNet,crm/tgrs22/RRNet,crm/tcsvt22/weaklySOD,crm/tcyb22/rsi,crm/tgrs19/rsi,zhang2019synthesizing}, and has been widely used in many related fields \cite{crm/tcsvt19/review}, such as object segmentation \cite{crm/ins21/superpixel}, content enhancement \cite{crm/JEI16/underwater,crm/tip21/underwaterMedium,crm/spl21/underwater,crm/cvpr20/low-light,crm/tmm20/dehazing,crm/tits22/low-light,crm/tmm22/blindSR,crm/acmmm21/bridgenet,crm/CVPR21/depthSR,crm/tip19/depthSR,crm/ijcai20/SR,crm/mtap22/dehazing}, and quality assessment \cite{crm/SPIC21/underwaterIQA,crm/tcsv22/underwaterIQA}. Chen \etal \cite{crm/aaai20/GCPANet} developed a method to make full use of global context. Liu \etal \cite{liu2018picanet} introduced a network to selectively attend to informative context locations for each pixel. In addition, the salient boundaries have been introduced into the model to improve the representation and highlight the desirable boundaries \cite{zhao2019egnet, qin2019basnet, wang2017edge}. Some methods  integrated features in multiple layers of CNN to exploit the context information at different semantic levels \cite{pang2020multi, wang2017edge}. In some challenging and complex single image scenarios, some works seek help from other modality data (\eg, depth map \cite{crm/acmmm21/CDINet,crm/tcyb21/ASIFNet,crm/tip21/DynamicRGBDSOD,crm/eccv20/RGBDSOD,crm/tmm22/3DSaliency,crm/tcyb20/going,crm/tip22/CIRNet} and thermal map \cite{crm/tmm22/TNet}). 
In addition, co-salient object detection (CoSOD) aims to detect salient objects from an image group containing several relevant images \cite{Zhang_2015_CVPR,crm/nips20/CoADNet,crm/tcyb22/glnet,crm/tmm19/HSCS,crm/ICME18/CoSOD,crm/tcyb19/iterativeCoSOD,crm/tip18/RGBDCoSOD,zhang2020adaptive, jiang2019unified,fan2021re}. 
The difference between CoSOD and VSOD is that it does not have temporal consistency, and the co-salient object is generally only consistent in semantic categories, rather than the same object.

\subsection{Salient Object Detection in Video}

The last decade has witnessed the considerable development of salient object detection in video sequences. Earlier VSOD methods mostly locate salient objects through hand-crafted features \cite{tu2017fusing,chen2017video,guo2017video,crm/tip19/VSOD}. 
Tu \etal \cite{tu2017fusing} detected the salient object in the video through two distinctive object detectors and refined the final spatiotemporal saliency result by measuring the foreground connectivity between two maps from two detectors. 
Chen \etal \cite{chen2017video} divided the long-term video sequence into some short batches and proposed to detect saliency in a batch-wise way, where the low-rank coherency is introduced to guarantee temporal smoothness.
However, the performance of these methods is not satisfactory due to the limited feature representation capabilities.
Recently, deep learning has demonstrated its power in VSOD tasks. 
Among them, some VSOD models adopt a single-stream structure that directly feeds the video sequences recursively into the network. 
For instance, 
Wang \etal \cite{wang2017video} proposed the first work applying deep learning to the VSOD task.
Li \etal \cite{li2018flow} proposed a two-stage FCN-based model, where the first stage is responsible for detecting static saliency, and the second stage is utilized to detect spatiotemporal saliency with two consecutive frames. 
In general, this method models saliency in a relatively primitive way. 
With the development of the model, some subtle module designs are proposed. 
For example, Song \etal \cite{song2018pyramid} used the designed Pyramid Dilated Bidirectional ConvLSTM to achieve deeper spatiotemporal feature extraction. 
Fan \etal \cite{fan2019shifting} introduced a VSOD model based on ConvLSTM, which is applied to model spatiotemporal features in a fixed length of video frames. Moreover, a new VSOD dataset with human visual fixation to model the human saliency shifting is proposed as well. 
Chen \etal \cite{chen2021novel} focused on the results derived from previous SOTA models, which are applied as pseudo labels to fine-tune a new model, considering the motion quality. 
Chen \etal \cite{chen2021exploring} presented a novel spatiotemporal modeling unit based on 3D convolution. 

In addition, another typical VSOD pipeline is the two-stream structure, where the optical flow image generated by FlowNet2 \cite{ilg2017flownet} or other methods is directly fed into the network as another stream input. 
Current two-stream models can be divided into two categories. One is the uni-direction guidance model, as shown in Fig. \ref{fig1}(a). 
Li \etal \cite{li2019motion} used the two-stream model to extract two modality features, where the temporal branch is designed to affect the spatial branch for better salient results. 
Ren \etal \cite{DBLP:conf/eccv/RenH0HH20} proposed to excite the video saliency branch with encoded optical flow features, and developed the semi-curriculum learning manner to learn saliency. 
The other is the undifferentiated and bidirectional fusion model, as shown in Fig. \ref{fig1}(b). 
Ji \etal \cite{ji2021full} proposed to exploit the cross-modality features by considering a mutual restraint scheme. 
Chen \etal \cite{chen2021confidence} tried to adaptively fuse features from motion and appearance via estimating confidence scores. 
Su \etal \cite{su2020ds} dynamically learned the weight vector of two modality features and aggregated the corresponding features complementarily. Zhao \etal \cite{Zhao_2021_CVPR} used scribble labels to train the VSOD model, in which cross-modality fusion and temporal constraint are used to model spatiotemporal information.

It is worth mentioning that the VSOD task is highly related to the unsupervised video object segmentation (VOS) task. Lu \etal \cite{lu2019see} proposed an unsupervised video object segmentation method, where a co-attention layer learns discriminative foreground information in video frame pairs. Zhou \etal \cite{DBLP:journals/tip/ZhouLWTS20} used an asymmetric motion-attentive transition to identify moving motion information and facilitate the representation of spatiotemporal cues in the zero-shot video object segmentation task. Wang \etal \cite{DBLP:conf/iccv/WangLSC019} built a fully connected graph to explore more representative and high-order relation information for zero-shot VOS. Cho \etal \cite{cho2022treating} regarded motion cues as optional in the unsupervised video object segmentation network, thereby designing a motion branch that can be adaptively turned on or off to participate or not in saliency detection.

The differences between our method and existing methods can be summarized in two major points. The previous two-stream structure is mainly a unidirectional guidance model or undifferentiated and bidirectional fusion model. But they may lead to insufficient information extraction due to wrong selection or ignoring the primary and secondary roles of different modalities (\ie, appearance and motion). Hence, we propose a more comprehensive and more secure strategy for modeling cross-modality interaction in the VSOD task under the two-stream structure, including an appearance-dominated branch and a motion-dominated branch. Two branches each consider the fusion with opposite modality tendencies, owning a clear and specific modality guidance tendency. Furthermore, to implement our overall framework, we also design concrete models that differ from existing methods, where the GDR module, CRC module, and IPF module cooperate to fully mobilize the relationship between different modalities and different detection branches.

\section{Methodology}
\subsection{Overview of Proposed Network}
As shown in Fig. \ref{fig2}, the proposed PSNet is a two-stream encoder-decoder network, following an up-down mirror-symmetrical structure. For the concise of the following description, we denote the current RGB frame as $\mathit{R}_{t}$, and the next RGB frame as $\mathit{R}_{t+1}$. These two adjacent images are input into FlowNet2 \cite{ilg2017flownet} to predict optical flow $\mathit{O}_{t,t+1}$ in an end-to-end way. 
With these inputs, the $\mathit{R}_{t}$ and $\mathit{O}_{t,t+1}$ are fed into 
the pre-trained ResNet50 backbone network that removes the last average pooling layer and the fully connected layer to obtain the encoder features of $\mathit{f}_{i}^{a}$ and $\mathit{f}_{i}^{m}$, where $i=\{1,2,3,4,5\}$ indicates the $i^{th}$ layer. 
The parameters of the spatial encoder and temporal encoder are not shared in our model.
In this network, we only use the features from the last four layers for the savings of computational costs. After that, both $\mathit{f}_{i}^{a}$ and $\mathit{f}_{i}^{m}$ are further input to the appearance-dominated branch and motion-dominated branch for feature decoding. As for the feature decoding process, we briefly illustrate the appearance-dominated branch as an example. First, all the dominant encoder features $\mathit{f}_{i}^{a}$ from the last four layers (\ie, $i=\{1,2,3,4,5\}$) are embedded into the GDR module to achieve the dominate-modality feature reinforcement and generate the corresponding reinforced dominant features $\mathit{f}_{i}^{a,r}$. Following that, the reinforced dominant features $\mathit{f}_{i}^{a,r}$, the corresponding appearance and motion features of $\mathit{f}_{i}^{a}$ and $\mathit{f}_{i}^{m}$, and the previous decoder features $\mathit{f}_{i+1}^{a,d}$ are input to the CRC module, thereby completing the explicit cross-modality information interaction and obtaining the decoder features $\mathit{f}_{i}^{a,d}$  of the current layer. Finally, we aggregate the outputs of the two decoder branches and generate the final saliency map through the IPF module. 

\subsection{Gather Diffusion Reinforcement Module}
\begin{figure}[t]
 	\centering
 	\includegraphics[width=\columnwidth]{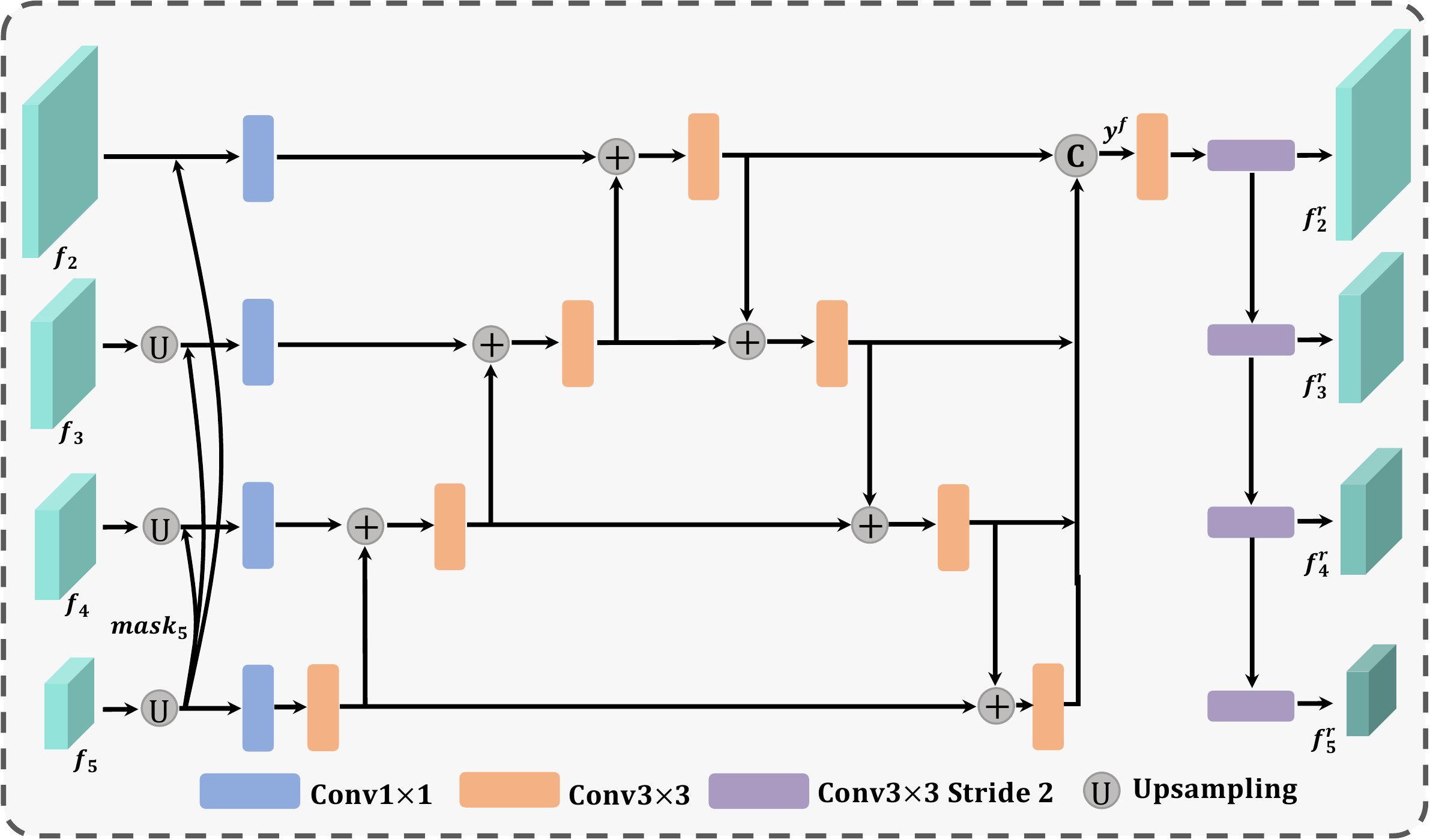}
 	\caption {The architecture of our proposed Gather Diffusion Reinforcement (GDR) module. 
 	}
 	\label{fig3}
 \end{figure}

As mentioned earlier, each of our decoding branches has clear dominant and auxiliary modality partitions. In order to ensure the effectiveness and comprehensiveness of the dominant modality features as much as possible, we consider the following motivations for designing a GDR module to strengthen the dominant features of each layer. In the encoding stage, the features extracted by each layer are relatively independent and have their own characteristics. With the network going deeper, the high-level features may contain more location and abstract semantic information about the salient object. At the same time, the low-level features are prone to have more detailed information, such as textures and boundaries. Both high-level and low-level features are essential for salient object detection and integrating them can help to generate high-quality multi-level features. Based on this, the primary function of our GDR module is to correlate the relationship between encoder features at different scales to develop more comprehensive encoder features. The detailed architecture of the GDR module is shown in Fig. \ref{fig3}.

\indent Given several features from different levels of the dominant branch, a Gather module is designed to exploit cross-layer and cross-scale information interaction. Specifically, considering that the features of different layers may contain some noise, especially in the low-level features, the coarse semantic mask predicted by the top encoder layer is used to filter out such noise, which is defined as:
\begin{align}
mask_{5} &=\sigma\left(\mathcal{C}_{3 \times 3}\left(\mathcal{C}_{3 \times 3}\left(f_{5}\right)\right)\right), \\
f_{i}^{s} &= Up\left(mask_{5}\right) \otimes f_{i},
\end{align}
where $\mathcal{C}_{3 \times 3}$ is convolution layer with the kernel size of $3 \times 3$, $\sigma$ denotes sigmoid function, $\otimes$ refers to element-wise multiplication operation, $f_{i}^{s}$ are the features after the semantic filtering, and $Up$ is the upsampling operation. For simplicity, the superscript $a$ or $m$ of the encoder features $f_{i}$ indicating the appearance branch or the motion branch is omitted. \\
\indent Subsequently, inspired by \cite{hong2021deep}, we use a recursive bidirectional structure to fuse multi-level features and establish relationships between features across scales in a hierarchical manner. That is, the fusion process is not limited to top-down but also explores bottom-up fusion to achieve more comprehensive multi-scale fusion. The top-down multi-scale feature interaction can be described as:
\begin{equation}
y_{i} = \left\{\begin{array}{cc}
\mathcal{C}_{3 \times 3}\left(\mathcal{C}_{1 \times 1}\left(f_{i}^{s}\right)+y_{i+1}\right),  i = \{2,3,4\} \\
\mathcal{C}_{3 \times 3}\left(\mathcal{C}_{1 \times 1}\left(f_{i}^{s}\right)\right),  \qquad \quad i  = 5
\end{array}\right.
\end{equation}
where $\mathcal{C}_{1 \times 1}$ is convolution layer with the kernel size of $1 \times 1$.

\indent Then, the reverse operation is also performed to achieve a more comprehensive cross-scale interaction:
\begin{equation}
y_{i}^{\prime}=\left\{\begin{array}{cc}
\mathcal{C}_{3 \times 3}\left(y_{i}+y_{i-1}^{\prime}\right), & i  =\{3,4,5\} \\
\mathcal{C}_{3 \times 3}\left(y_{i}\right), &  i=2
\end{array}\right.
\end{equation}
\indent As such, all interaction features $\left\{y_{i}^{\prime} \mid i=\{2,3,4,5\}\right\}$ are fused together in the form of concatenation-convolution: 
\begin{equation}
    y^{f}=\mathcal{C}_{3 \times 3}\left(\operatorname{Cat}\left[y_{2}^{\prime}, y_{3}^{\prime}, y_{4}^{\prime}, y_{5}^{\prime}\right]\right),
\end{equation}
where $\operatorname{Cat}$ is channel-wise concatenation operation. \\
\indent Finally, considering that each level of CRC needs a different scale of features from GDR, a straightforward way is to use a diffusion module to diffuse features. Here, we perform 3×3 convolution with stride 2 on the fusion features and generate the reinforced features:
\begin{equation}
    f_{i}^{r}=\left\{\begin{array}{c}
\mathcal{C}_{3 \times 3 \operatorname{stride} 2}\left(y^{f}\right), \quad i=2 \\
\mathcal{C}_{3 \times 3 \operatorname{stride} 2}\left(f_{i-1}^{r}\right), \quad i  =\{3,4,5\}
\end{array}\right.
\end{equation}
where $\mathcal{C}_{3 \times 3 \operatorname{stride}2}$ denotes $3 \times 3$ convolution with the stride of 2. The reinforced features in the appearance-dominated branch and motion-dominated branch can be distinguished as $\mathit{f}_{i}^{a,r}$ and $\mathit{f}_{i}^{m,r}$, respectively. 
In fact, both DSS \cite{DBLP:journals/pami/HouCHBTT19} and our GDR module adopt a structure similar to FPN \cite{DBLP:conf/cvpr/LinDGHHB17}, which is used for enriching the representation of multi-scale information. But our GDR module acts as a single-modality feature enhancement with cross-level, cross-scale information, and then passes them to the decoder. In the implementation, for the FPN and short connection in \cite{DBLP:journals/pami/HouCHBTT19}, the multi-scale information is fused in a single direction (up-to-down). While for our GDR module, the interaction direction is not restricted to a single direction but follows a recursive way to achieve more comprehensive multi-scale information. In addition, the high-level features from the encoder are used to filter out the noise in low-level features for a more robust single-modality representation in our GDR module.

 \begin{figure}[t]
 	\centering
 	\includegraphics[width=\columnwidth]{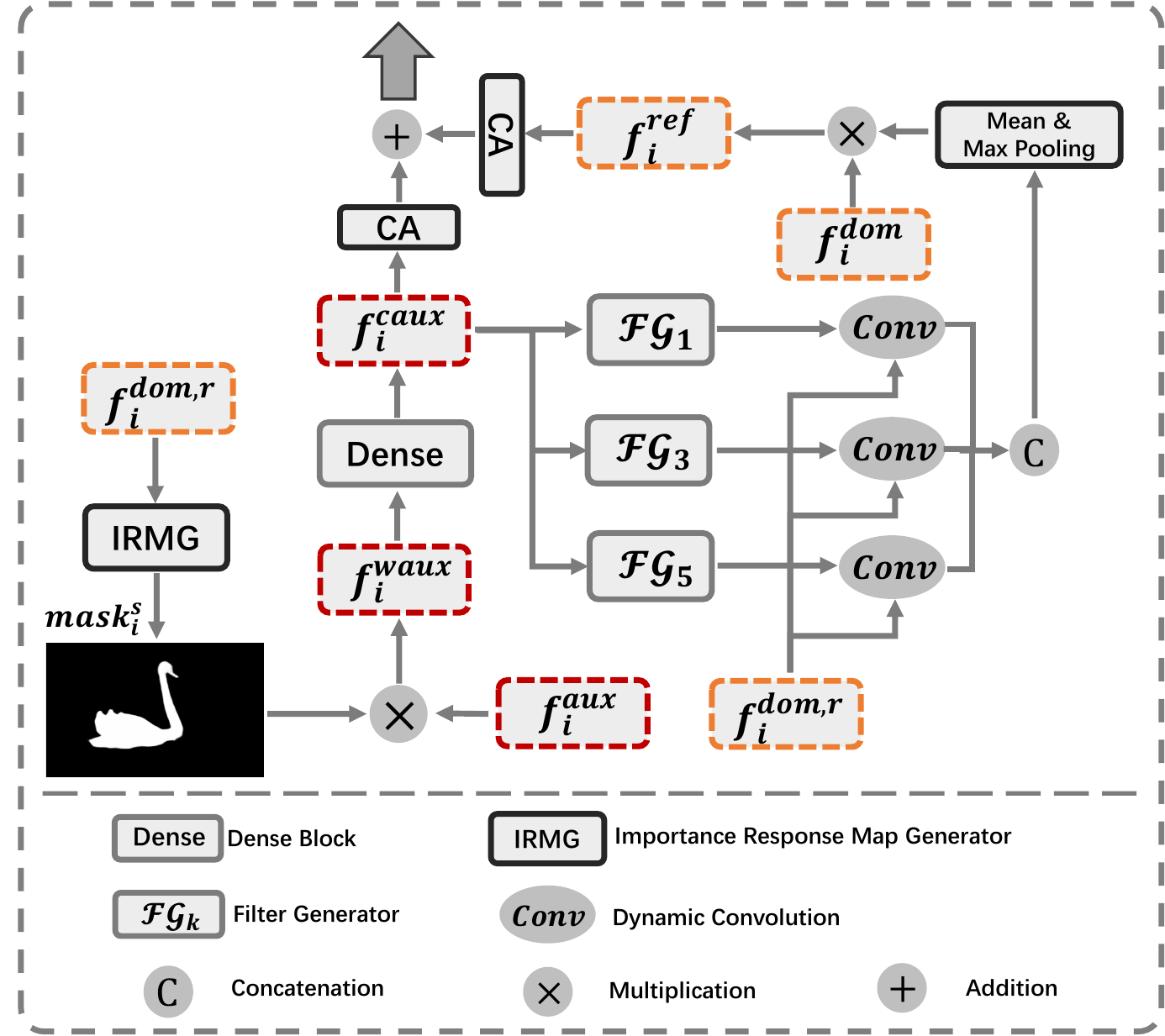}
 	\caption {The structure of our proposed Cross-modality Refinement and Complement (CRC) module. 
 	}
 	\label{fig4}
 \end{figure}

\subsection{Cross-modality Refinement and Complement Module}
The cross-modality interaction has always been a core topic in video salient object detection. Fortunately, under our parallel symmetric architecture, each branch has a definite dominant modality and a corresponding auxiliary modality, so that we can design the interaction module more explicitly and clearly. For concise, we denote the dominant modality features as $f_{i}^{dom}$, and the auxiliary features as $f_{i}^{aux}$. 
The structure of CRC is depicted in Fig. \ref{fig4}. As mentioned earlier, for each dominated branch, the $f_{i}^{dom}$ are more dominant than $f_{i}^{aux}$, but this does not mean that auxiliary features are entirely useless. In other words, there is still some helpful information in $f_{i}^{aux}$, which will contribute to the saliency feature learning. Therefore, starting from the auxiliary modality, we divide its role into two types of refinement and complement and design a CRC module to maximize the use of auxiliary information. Furthermore, the features of $f_{i}^{dom}$ are fed into the GDR module to generate the reinforced dominant features $f_{i}^{dom,r}$ for the current CRC module.

On the one hand, the supplement of the feature dimension is the most direct from the perspective of information interaction. But direct and indiscriminate integration of  $f_{i}^{aux}$ may introduce contamination noise into the dominant features. Therefore, we try to select auxiliary features from the perspective of dominant features and determine the feature components that need to be supplemented. Specifically, two convolutions are employed on the $f_{i}^{dom,r}$ to obtain the importance response map $mask_{i}^{s}$. Then, we use this map to weight the auxiliary features of $f_{i}^{aux}$ and utilize the dense block and residual block to strengthen the features, thereby determining the auxiliary features that need to be supplemented to the reinforced dominant features. The above operations are presented as follows:
\begin{align}
&{mask}_{i}^{s}=\sigma\left(\mathcal{C}_{1 \times 1}\left(\mathcal{C}_{3 \times 3}\left(f_{i}^{{dom,r }}\right)\right)\right), \\
&f_{i}^{ {waux }}={mask}_{i}^{s} \otimes f_{i}^{a u x}, \\
&f_{i}^{{caux }}=\mathcal{C}_{1 \times 1}\left({Dense}\left(f_{i}^{ {waux }}\right)+f_{i}^{ {waux }}\right),
\end{align}
where $f_{i}^{ {waux }}$ are the weighted auxiliary features, $f_{i}^{ {caux }}$ denote the final complemented auxiliary features, and $Dense$ represents the dense block in \cite{huang2017densely}.

\indent On the other hand, in addition to the complement of feature dimensions, auxiliary features can also be used to refine the irrelevant interference and misinformation in the dominant features. 
However, considering the difference and interference noise of the two modalities, as well as the large variation of the characteristics of the two modalities with the scene, we do not directly apply $f_{i}^{caux}$ to refine $f_{i}^{dom,r}$, but introduce the dynamic convolution filters \cite{jia2016dynamic, he2019dynamic} to adaptively generate the convolution kernel for different scenarios, so as to ensure the generalization and robustness of the network. With the complemented auxiliary features $f_{i}^{caux}$, we use the local dynamic convolution \cite{jia2016dynamic} with different dilated rates to generate multi-scale convolution kernels. 
Then, the generated dynamic convolution kernels are used to convolve the reinforced dominant features $f_{i}^{dom,r}$, achieving the goal of refining the details. The process can be expressed by:
\begin{align}
k_{1} &=\mathcal{F} \mathcal{G}_{1}\left(f_{i}^{caux }\right) \circledast  f_{i}^{dom,r}, \\
k_{3} &=\mathcal{F} \mathcal{G}_{3}\left(f_{i}^{caux }\right) \circledast  f_{i}^{dom,r}, \\
k_{5} &=\mathcal{F} \mathcal{G}_{5}\left(f_{i}^{caux }\right) \circledast  f_{i}^{dom,r}, \\
f_{i}^{d y} &=\mathcal{C}_{3 \times 3}\left(\operatorname{Cat}\left[k_{1}, k_{3}, k_{5}\right]\right),
\end{align}
where $\mathcal{F} \mathcal{G}_{j}$ presents the filter generator with the dilated rate of $j$ by using two convolutions and reshaping operations, and $\circledast$ indicates convolution operation. 

\indent Next, we employ $f_{i}^{dy}$ to generate a refinement mask, and then revise and refine the features of $f_{i}^{dom}$:
\begin{align}
{C}_{i} = \operatorname{Cat}&\left({maxpool}\left(f_{i}^{d y}\right),  {avgpool}\left(f_{i}^{d y}\right)\right),\\ 
&{mask}_{i}^{r}=\sigma\left(\mathcal{C}_{3 \times 3}\left({C}_{i}\right)\right), \\
&f_{i}^{ref}={mask}_{i}^{r} \otimes f_{i}^{ {dom}},
\end{align}
where $f_{i}^{ref}$ denote the refined dominant features, ${mask}_{i}^{r}$ is the generated refinement mask, and $maxpool$ and $avgpool$ denote max-pooling and average pooling respectively. 
\indent Finally, we combine the complemented auxiliary features $f_{i}^{caux}$ and refined dominant features $f_{i}^{ref}$ after the channel compaction:
\begin{equation}
    f_{i}^{rc}=C A\left(f_{i}^{caux}\right)+C A\left(f_{i}^{ref}\right),
\end{equation}
where $CA$ presents channel attention block \cite{hu2018squeeze,crm/tce22/covid,crm/tim22/covid}.

\indent With the final interaction features $f_{i}^{rc}$, we combine them with the decoder features of the previous layer $f_{i+1}^{dom,d}$ to generate the final decoder features of the current layer:
\begin{equation}
    f_{i}^{dom,d}=\left\{\begin{array}{c}
\mathcal{C}_{3 \times 3}\left(\operatorname{Cat}\left(f_{i}^{rc}, U p\left(f_{i+1}^{dom,d}\right)\right)\right), i=\{2,3,4\} \\
\mathcal{C}_{3 \times 3}\left(\operatorname{Cat}\left(f_{i}^{rc}, f_{i}^{dom,r}\right)\right), i=5 
\end{array}\right.
\end{equation}
where the decoder features of $i^{th}$ layer in the appearance-dominated branch and motion-dominated branch are as $\mathit{f}_{i}^{a,d}$ and $\mathit{f}_{i}^{m,d}$, respectively.


 \begin{figure}[t]
 	\centering
 	\includegraphics[width=\columnwidth]{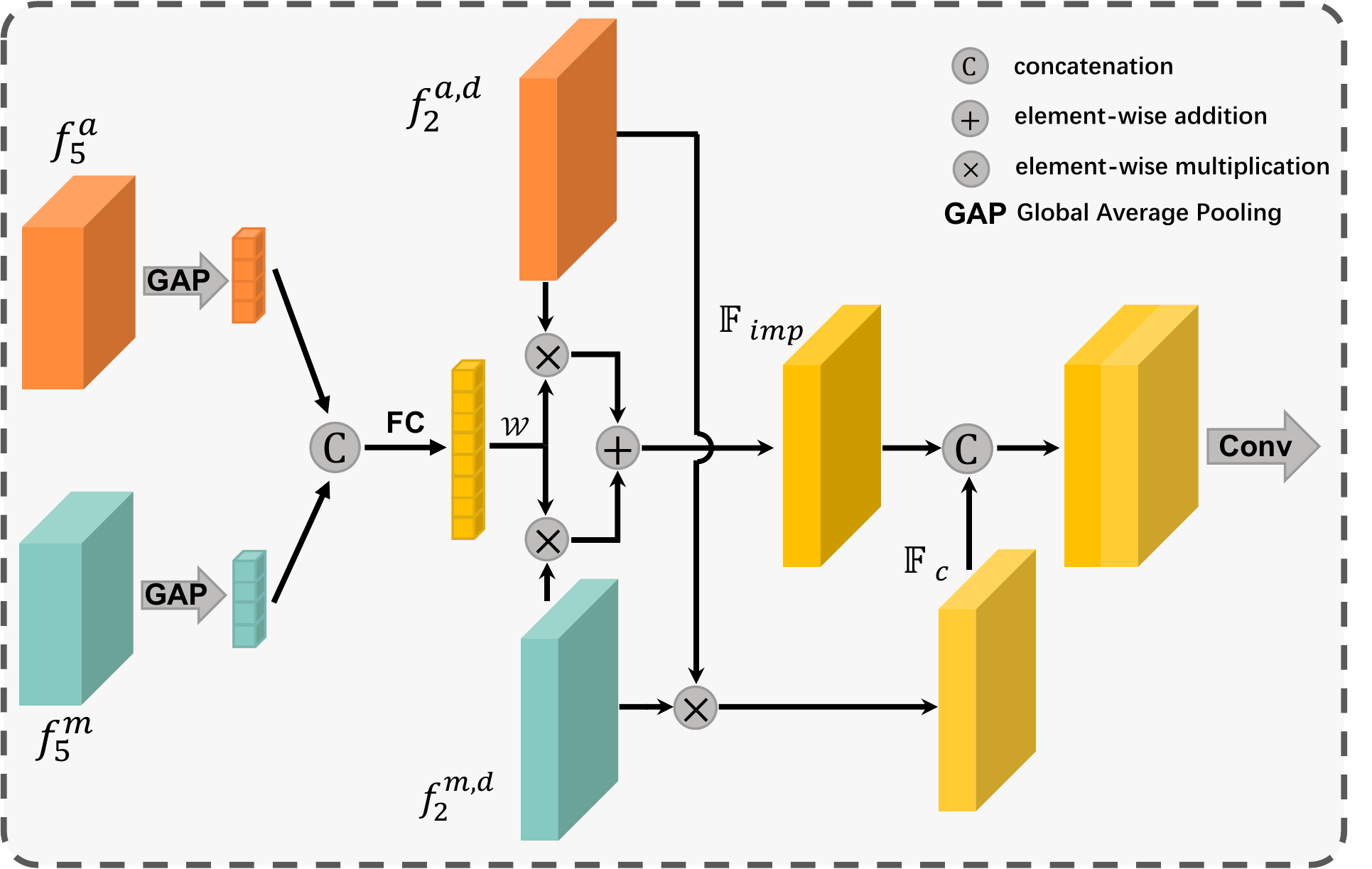}
 	\caption {The structure of our proposed Importance Perception Fusion (IPF) module. 
 	}
 	\label{fig5}
 \end{figure}
\subsection{Importance Perception Fusion Module}
Under the parallel symmetric framework, the appearance-dominated branch and motion-dominated branch generate comprehensive spatiotemporal features corresponding to different modality dominance with clear and well-defined roles. Although both branches can be regarded as complete VSOD branches, the dominant modality they set is different, and the learned features are also different.
To obtain a more robust and generalized final result, inspired by the MF module in \cite{crm/tip21/DPANet}, we introduce an IPF module to achieve the branch fusion, considering the different importance of the upper and lower branches in different scenarios.
Fig. \ref{fig5} illustrates the flowchart of the IPF module.  
The input features of the IPF module can be divided into two parts. One is the last-layer features of the encoder ($f_5$), which is used to perceive the different importance of the upper and lower branches in different scenes. 
In this way, an adaptive importance weight $\mathcal{W} \in \mathbb{R}^{128}$ is learned.
More specifically, the features from the $5^{th}$ layer of appearance and motion encoders (\ie, $f_{5}^{a}$ and $f_{5}^{m}$) are first fed into global average polling and then concatenated together to learn a channel-wise weight:
\begin{equation}
    \mathcal{W}=\sigma\left(F C\left(\operatorname{Cat}\left(G A P\left(f_{5}^{a}\right), G A P\left(f_{5}^{m}\right)\right)\right)\right),
\end{equation}
where $GAP$ is global average pooling, and $FC$ represents a full-connected layer. \\

Another input to the IPF module is the top-layer features of two decoders. The reason why we choose top-layer features $f_2^d$ of the decoder is that the $f_2^d$ has higher resolution than $f_3^d$ and $f_4^d$, and contains more comprehensive decoding information, which is more suitable for our purpose in IPF.
More specifically, we use the weight $\mathcal{W}$ to combine the output decoder features of two branches $f_{2}^{a,d}$ and $f_{2}^{m,d}$ into importance weighted features $\mathbb{F}_{i m p}$: 
\begin{equation}
    \mathbb{F}_{i m p}=\mathcal{W} \odot f_{2}^{a, d}+(1-\mathcal{W}) \odot f_{2}^{m,d},
\end{equation}
where $f_{2}^{a,d}$ and $f_{2}^{m,d}$ are the decoder features of last layer in the corresponding branch, and $\odot$ denotes element-wise multiplication with the broadcasting strategy. 

Furthermore, the common response between two outputs from two branches is also important for the final saliency result. Thus, a simple but effective way is to perform multiplication to highlight the common part of the two branches: 
\begin{equation}
    \mathbb{F}_{c}=f_{2}^{a,d} \otimes f_{2}^{m,d}.
\end{equation}
Finally, the common features $\mathbb{F}_{c}$ and importance weighted features $\mathbb{F}_{imp}$ are combined by concatenation operation to predict the final saliency map:
\begin{equation}
     { pre }_{s}=\sigma\left(\mathcal{C}_{3 \times 3}\left(\mathcal{C}_{3 \times 3}\left(\operatorname{Cat}\left(\mathbb{F}_{c}, \mathbb{F}_{i m p}\right)\right)\right)\right).
\end{equation}
\subsection{Loss Function}
The network is trained in a multiple supervision manner for the sake of faster convergence and better performance. First, for the final saliency results generated by the IPF module, we employ a joint loss function $\mathcal{L}_{{sal}} $ to train our model, which is given by:
\begin{equation}
\mathcal{L}_{{sal}}=\mathcal{L}_{ {bce }}\left({pre}_{s}, GT\right)
+\mathcal{L}_{ {ssim }}\left( { pre }_{s}, G T\right),
\end{equation}
where $\mathcal{L}_{bce} $ is the binary cross-entropy loss, and $\mathcal{L}_{ssim} $ is the structural similarity loss. 

In addition, we add the side-output supervision on each branch. Taking the appearance-dominated branch as an example, firstly, the saliency map $S_{a}$ predicted by the appearance-dominated branch will be trained by the $\mathcal{L}_{{sal}}$ loss. Besides, the intermediate saliency results from GDR and CRC modules are also supervised by the ground truth. Specifically, the supervisions include: (1) the appearance backbone saliency map ${mask}_{5}$ deduced from the $5^{th}$ layer of appearance backbone, which is employed in GDR as a noise filter; (2) the importance response map $ {mask}_{i}^{s}$ deduced from each CRC in the appearance-dominated branch, in which $i=\{2,3,4,5\}$. Therefore, for the appearance-dominated branch, the loss function can be formulated as:
\begin{equation}
    \begin{aligned}
\mathcal{L}_{{appearance }} &=\mathcal{L}_{ {sal }}\left(S_{a}, G T\right)+\lambda_{1} \mathcal{L}_{ {bce }}\left({mask}_{5}, G T\right) \\
&+\lambda_{2} \sum_{i=2}^{5} \mathcal{L}_{b c e}\left({mask}_{i}^{s}, G T\right)
\end{aligned},
\end{equation}
where $\lambda_1$ and $\lambda_2$ are hyper-parameters for balancing the losses, which are empirically set to 0.6 and 0.4, respectively. To fit the size of the predicted map, all ground truth will be downsampled to the size of the corresponding predicted map. Similarly, we can get the loss function of the motion-dominated branch, denoted as $\mathcal{L}_{motion} $. \\
\indent Finally, the total loss is defined as follows:
\begin{equation}
    \mathcal{L}_{{total}}=\mathcal{L}_{{sal }}+\mathcal{L}_{ {appearance }}+\mathcal{L}_{{motion }}.
\end{equation}

\begin{table*}[ht]
\renewcommand\arraystretch{1.06}
			\caption{
			Quantitative results on the DAVIS, SegV2, DAVSOD, and ViSal datasets. The top two score was marked in \textbf{bold} and \underline{underline}, respectively. 
			}
\vspace{-5 mm}
	\begin{center}
		\resizebox*{1\textwidth}{!}{
\begin{tabular}{|l|c|c|c|c|c|}
				\hline
				\multirow{2}{*}{\textbf{Methods}}&\multirow{2}{*}{\textbf{Years}}&\textbf{DAVIS} &\textbf{SegV2} &\textbf{ViSal} &\textbf{DAVSOD}\\
				\cline{3-6} & &\metrics&\metrics&\metrics&\metrics\\
				\hline
				\multicolumn{6}{|c|}{\textbf{Deep Learning Static Salient Obejct Detection}} \\
				\hline
				
EGNet \cite{zhao2019egnet}& 2019 &\triplets(0.768, 0.829, {0.056})&\triplets(0.774, 0.845, 0.024)&\triplets(0.941, 0.946 , 0.015)&\triplets(0.604, 0.719, 0.101)\\
CPD \cite{wu2019cascaded}& 2019 &\triplets(0.778, 0.859, {0.032})&\triplets(0.778, 0.841, 0.023)&\triplets(0.941, 0.942 , 0.016)&\triplets(0.608, 0.724, 0.092)\\
ITSD \cite{zhou2020interactive}& 2020 &\triplets(0.835, 0.876, {0.033})&\triplets(0.807, 0.787, 0.027)&\fillcell&\triplets(0.651, 0.747, 0.094)\\
\hline
				\multicolumn{6}{|c|}{\textbf{Traditional Video Salient Object Detection}} \\
				\hline
MSTM \cite{tu2016real}& 2016 &\triplets(0.429, 0.583, {0.165})&\triplets(0.526, 0.643, 0.114)&\triplets(0.673, 0.749, 0.095)&\triplets(0.344, 0.532, 0.211)\\
SGSP \cite{liu2016saliency}& 2017 &\triplets(0.655, 0.592, {0.138})&\triplets(0.673, 0.681, 0.124)&\triplets(0.677, 0.706, 0.165)&\triplets(0.426, 0.577, 0.207)\\
STBP \cite{xi2016salient}& 2016 &\triplets(0.544, 0.677, {0.096})&\triplets(0.640, 0.735, 0.061)&\triplets(0.622, 0.629, 0.163)&\triplets(0.410, 0.568, 0.160)\\
FDOS \cite{tu2017fusing} & 2017 &\triplets(0.701, 0.784, 0.061)&\triplets(0.683, 0.765, 0.045)&\triplets(0.767, 0.801, 0.063)&\triplets(0.456, 0.582, 0.157)\\
SCOM \cite{chen2018scom}& 2018 &\triplets(0.783, 0.832, 0.048)&\triplets(0.764, 0.815, 0.030)&\triplets(0.831, 0.762, 0.122)&\triplets(0.464, 0.599, 0.220)\\
SFLR \cite{chen2017video} & 2017 &\triplets(0.727, 0.790, 0.056)&\triplets(0.745, 0.804, 0.037)&\triplets(0.779, 0.814, 0.062)&\triplets(0.478, 0.624, 0.132)\\
\hline
				\multicolumn{6}{|c|}{\textbf{Deep Learning Video Salient Object Detection}} \\
				\hline
SCNN \cite{tang2018weakly}& 2018 &\triplets(0.714, 0.783, {0.064})&\fillcell&\triplets(0.831, 0.847, 0.071)&\triplets(0.532, 0.674, 0.128)\\
DLVS \cite{wang2017video}& 2018 &\triplets(0.708, 0.794, {0.061})&\fillcell&\triplets(0.852, 0.881, 0.048)&\triplets(0.521, 0.657, 0.129)\\
FGRN \cite{li2018flow}& 2018 &\triplets(0.783, 0.838, {0.043})&\fillcell&\triplets(0.848, 0.861, 0.045)&\triplets(0.573, 0.693, 0.098)\\

MBNM  \cite{li2018unsupervised}& 2018 &\triplets(0.861, 0.887, {0.031})&\triplets(0.716, 0.809, {0.026})&\triplets(0.883, 0.898, 0.020)&\triplets(0.520, 0.637, 0.159)\\
PDB \cite{song2018pyramid}& 2018 &\triplets(0.861, 0.887, {0.028})&\triplets(0.800, 0.864, {0.024})&\triplets(0.888, 0.907, 0.032)&\triplets(0.572, 0.698, 0.116)\\
RCR  \cite{yan2019semi}& 2019 &\triplets(0.855, 0.882, {0.027})&\triplets(0.781, 0.842, {0.035})&\triplets(0.906, 0.922, 0.026)&\triplets(0.653, 0.741, 0.087)\\
SSAV  \cite{fan2019shifting}& 2019 &\triplets(0.861, 0.893, {0.026})&\triplets(0.801, 0.851, {0.023})&\triplets(0.939, 0.943, 0.020)&\triplets(0.603, 0.724, 0.092)\\
 MGA \cite{li2019motion}& 2019 &\triplets(0.892, 0.910, {0.023})&\triplets(0.821, 0.865, {0.030})&\triplets(0.933, 0.936, 0.017)&\triplets(0.640, 0.738, 0.084)\\
PCSA \cite{gu2020pyramid}& 2020 &\triplets(0.880, 0.902, {0.022})&\triplets(0.810, 0.865, {0.025})&\triplets(0.940, 0.946, 0.017)&\triplets(0.655, 0.741, 0.086)\\
CASNet \cite{ji2020casnet}& 2020 &\triplets(0.860, 0.873, {0.032})&\triplets(0.847, 0.820, {0.029}) &\fillcell&\fillcell\\
DSNet  \cite{su2020ds}& 2020 &\triplets(0.891, 0.914, \underline{0.018})&\triplets(0.832, 0.875, 0.028)&\triplets({0.950}, 0.949, \underline{0.013})&\fillcell\\
STVS \cite{chen2021exploring}& 2021 &\triplets(0.865, 0.892, \underline{0.018})&\triplets(\textbf{0.860}, \textbf{0.891}, \underline{0.017})&\triplets(\underline{0.952}, \underline{0.952}, \underline{0.013})&\triplets(0.651, 0.746, 0.086)\\
WVSOD \cite{Zhao_2021_CVPR} & 2021 &\triplets(0.793, 0.846, 0.038)&\triplets(0.762, 0.819, 0.033)&\triplets(0.875, 0.883, 0.035)&\triplets(0.593, 0.694, 0.115)\\
TransVOS  \cite{mei2021transvos}& 2021 &\triplets(0.869, 0.885, \underline{0.018})&\triplets(0.800, 0.816, {0.024})&\triplets(0.928, 0.917, 0.021)&\fillcell\\
CAG \cite{chen2021confidence}& 2021 &\triplets({0.898}, 0.906, \underline{0.018})&\triplets(0.826, 0.865, {0.027})&\triplets({0.950}, {0.950}, \underline{0.013})&\triplets({0.670}, {0.762}, \textbf{0.072})\\
FSNet \cite{ji2021full} & 2021 &\triplets(\textbf{0.907}, \textbf{0.920}, 0.020)&\triplets(0.805, 0.870, 0.024)&\fillcell&\triplets(\textbf{0.685}, \textbf{0.773}, \textbf{0.072})\\
PSNet  & - &\triplets(\textbf{0.907}, \underline{0.919}, \textbf{0.016})&\triplets(\underline{0.852}, \underline{0.889}, \textbf{0.016})&\triplets(\textbf{0.955}, \textbf{0.954}, \textbf{0.012})&\triplets(\underline{0.678}, \underline{0.765}, {0.074}) \\
\hline
			\end{tabular}}
			\label{tab:pf2}
		\end{center}
\end{table*}
\section{Experiments}
\subsection{Datasets and Evaluation Metrics}
We conduct experiments on four widely used public VSOD datasets in order to fully evaluate the effectiveness of our proposed method, \ie, DAVIS\cite{perazzi2016benchmark}, DAVSOD\cite{fan2019shifting}, SegV2\cite{li2013video}, and ViSal\cite{li2017benchmark}. \textbf{DAVIS} \cite{perazzi2016benchmark} dataset consists of $50$ clips of 480p and 720p videos with high-quality dense annotations, which is further split into $30$ videos for training and $20$ videos for testing. \textbf{DAVSOD} \cite{fan2019shifting} dataset includes $226$ clips of densely annotated videos, where the salient objects are annotated by dynamic eye-tracking. In DAVSOD, $80$ clips of videos are for testing. \textbf{SegV2} \cite{li2013video} dataset is an early proposed dataset with $14$ videos and $1065$ annotated frames, including multiple objects that make it more challenging than others. \textbf{ViSal} \cite{li2017benchmark} is a dataset containing $19$ videos with $193$ pixel-wise annotated frames. In this paper, the test will be carried out on the whole datasets of ViSal and SegV2, and the testing subset of DAVIS and DAVSOD datasets.
To quantitatively evaluate the effectiveness of the proposed method, we introduce three evaluation metrics, including maximum F-measure ($F_{\beta}$) \cite{achanta2009frequency}, S-measure ($S_{m}$) \cite{fan2017structure}, and Mean Absolute Error (MAE) \cite{crm/nc20/rsi}. 
For these three metrics, except for the MAE score, larger values of the F-measure and S-measure indicate better performance.

\subsection{Implementation Details}
We use the Pytorch toolbox to implement our network and train our model with an NVIDIA GTX3090 GPU.  We also implement our network by using the MindSpore Lite tool\footnote{\url{https://www.mindspore.cn/}}. Following the setting in \cite{ji2021full}, we use the stage-wise training protocol with image saliency datasets and video saliency datasets to train our model. In the first stage, we initialize our spatial backbone with a ResNet-50\cite{he2016deep}. Following \cite{ji2021full}, we remove the CRC and IPF modules, and pre-train this model on the training set of the DUTS dataset\cite{wang2017learning}. 
In this stage, the batch size and initial learning rate are set to $16$ and $0.002$, respectively. Moreover, the learning rate decays $0.1$ times per $10$ epochs. 
In the second stage, we use FlowNet2\cite{ilg2017flownet} to generate the corresponding optical flow map for each frame of the DAVIS dataset and pre-train the temporal branch. The training settings are the same as stage 1. Next, in stage 3, we use the DAVIS dataset, including RGB images and optical flow maps, to fine-tune our whole PSNet. Concretely, we load the learned weights from stage 1 and stage 2 to the spatial branch and temporal branch, respectively. The number of batch sizes is set to $8$. The learning rate is set to $0.0002$ for finer learning and stops learning after $20$ epochs. 
In each stage, we use the stochastic gradient descent (SGD) optimizer to train our model with a momentum of $0.9$ and a weight decay of $0.0005$. We resize all input images to $384 \times 384$. Furthermore, we apply a multi-scale training strategy with scales of $\{0.75,1,1.25\}$, random horizontal flipping, and random vertical flipping to enhance the generalizability and stability of our trained model.

\subsection{Comparison with the State-of-the-arts}
Our proposed method is compared with 25 state-of-the-art methods, including three static SOD methods (EGNet\cite{zhao2019egnet}, CPD\cite{wu2019cascaded}, ITSD\cite{zhou2020interactive}), six traditional VSOD methods (MSTM\cite{tu2016real}, STBP\cite{xi2016salient}, SGSP\cite{liu2016saliency}, SCOM\cite{chen2018scom}, FDOS\cite{tu2017fusing}, SFLR\cite{chen2017video}), and sixteen deep learning-based VSOD methods (SCNN\cite{tang2018weakly}, DLVS\cite{wang2017video}, FGRN\cite{li2018flow}, MBNM\cite{li2018unsupervised}, PDB\cite{song2018pyramid},  RCR\cite{yan2019semi}, SSAV\cite{fan2019shifting}, MGA\cite{li2019motion}, PCSA\cite{gu2020pyramid}, CASNet\cite{ji2020casnet}, WVSOD\cite{Zhao_2021_CVPR}, DSNet\cite{su2020ds}, TransVOS\footnote{TransVOS is a semi-supervised VOS method.}\cite{mei2021transvos}, STVS\cite{chen2021exploring}, CAG\cite{chen2021confidence}, FSNet\cite{ji2021full}). For fair comparisons, all the saliency maps are provided by authors or tested by the released code under the default parameters.

\begin{figure*}[!t]
	\centering
	\includegraphics[width=1\textwidth]{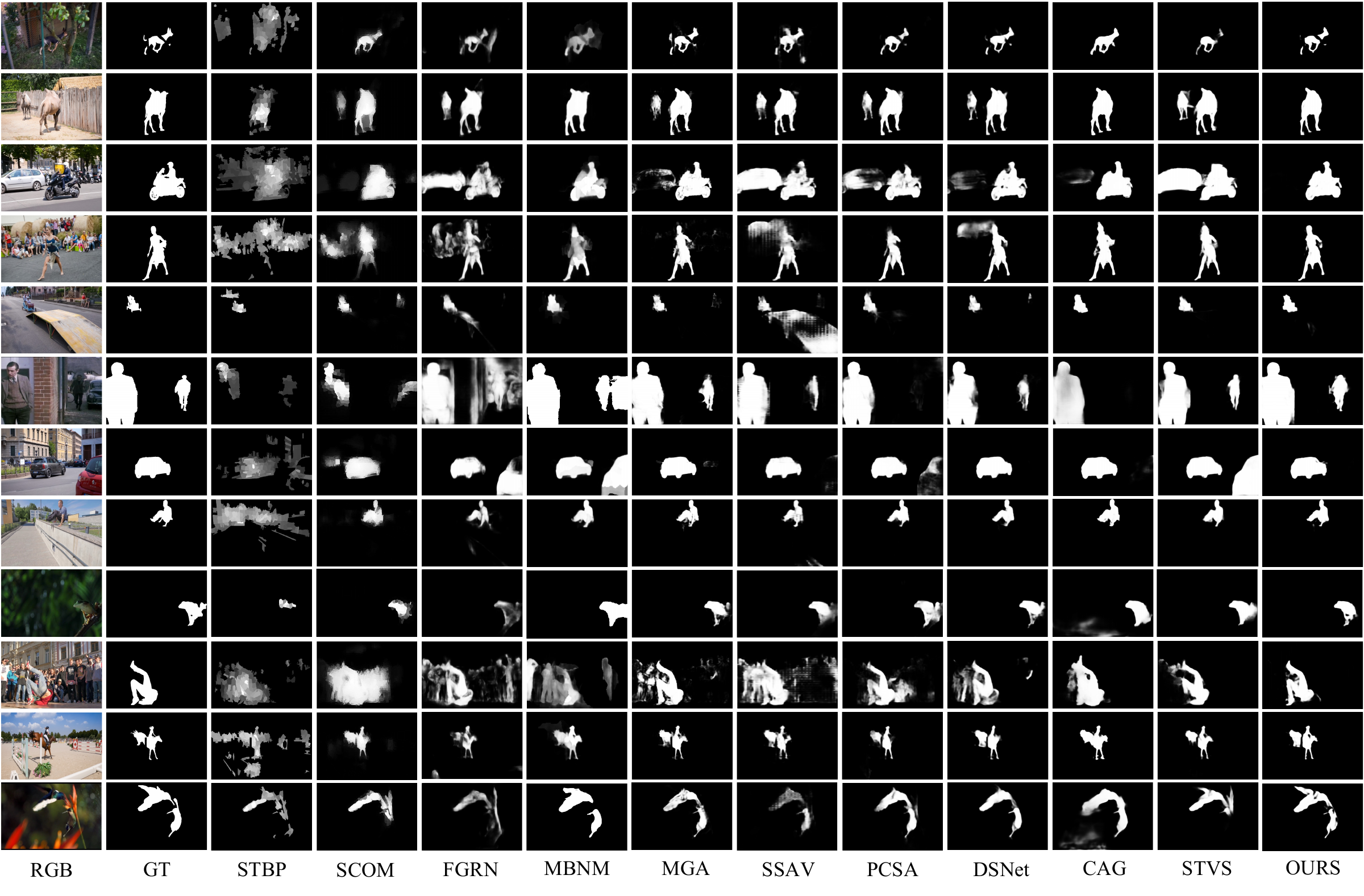}
	\caption {The visualization results of different video salient object detection methods. }
	\label{fig6} 
\end{figure*}

\subsubsection{\textbf{Quantitative Evaluation}}
The quantitative results max-F, S-measure, and MAE results are listed in Table \ref{tab:pf2}. The static SOD methods perform well in some simple datasets (\eg, ViSal) and even outperform VSOD methods, mainly because the image appearance cues in these datasets dominate most scenes.
Quantitatively, the static SOD method CPD\cite{wu2019cascaded} wins the percentage gain of 0.9\% in $F_{\beta}$ and 31.3\% in MAE against the MGA\cite{li2019motion} method on the ViSal dataset. 
However, this advantage will no longer exist in the face of complex video scenes, such as the DAVSOD dataset, whose performance is far lower than the VSOD methods.
For traditional VSOD methods, due to the limitation of only using hand-crafted features, their performance is even lower than that of static SOD methods. Taking the best traditional VSOD method SCOM\cite{chen2018scom} as an example, it achieves comparable performance with some deep learning-based static SOD methods (\eg, EGNet\cite{zhao2019egnet} and CPD\cite{wu2019cascaded}) on the DAVIS dataset. However, its performance is 50\% lower than deep learning-based static SOD methods on the DAVSOD dataset. 
Seeing Table \ref{tab:pf2}, it is observed that our method achieves competitive performance on these four datasets, basically ranking in the top two. 
Specifically, our method outperforms all other models on the ViSal dataset, which achieves the percentage gain of 7.6\% in terms of MAE score compared with the \textbf{second best} method (\ie, CAG\cite{chen2021confidence}).
In addition, compared with the \textbf{second best} model on the DAVIS dataset, the percentage gain reaches 11.1\% for the MAE score. 
Our method is slightly inferior to the FSNet method \cite{ji2021full} on the DAVSOD dataset, but achieves comparable performance on the DAVIS dataset, and has a clear performance advantage on the SegV2 dataset. From the analysis of the model size, the size of our method (67.9 M) is only about 80\% of the size of FSNet (83.4 M). Overall, our method still has certain advantages in terms of performance and model size.

The training time of PSNet is about 20 hours for all three stages of training, and the testing speed of our PSNet reaches 19 FPS with the model size of 67.9 M. 
Compared with optical-flow-based methods, such as the MGA (47 FPS and 91.5 M)\cite{li2019motion} and CAG (29 FPS and 55.3 M)\cite{chen2021confidence}, although the performance of our algorithm achieves the best result, our testing time and model size are not optimal, which is related to the inclusion of operations such as dynamic convolution in our network design. That is, at present, our model is still far from the real-time effect. Therefore, in the future, we can consider lightweight alternative modules to further improve the efficiency of model testing.

\subsubsection{\textbf{Qualitative Evaluation}}
To further illustrate the advantages of our proposed method, we provide some qualitative saliency results in Fig. \ref{fig6}. Compared with other methods, our method achieves superior results with complete object structure, precise saliency location, and sharp boundaries.
As can be seen, the traditional VSOD methods cannot achieve desirable results due to their limitations and deficiencies in feature representation, such as the $3^{rd}$ and $4^{th}$ columns. By contrast, deep learning-based methods achieve more competitive results, especially our proposed method is capable of addressing scenes with small objects and complex backgrounds. Taking the $2^{nd}$ and $3^{rd}$ rows as an example, these two sequences contain some tough challenges, in which the background exists some moving but non-salient objects. However, most methods, such as STVS \cite{chen2021exploring} and PCSA \cite{gu2020pyramid}, cannot completely suppress the distracting background objects in such complicated scenes. Thanks to the design of our network, our model can completely suppress such background disturbances that consider motion modality as the dominant feature in such scenes and reduce the interference of wrong appearance cues.
Meanwhile, in the $10^{th}$ row, the scene is more complex, where the man dancing in front of the audience is our salient object. 
However, the motion of the foreground objects changes very quickly, and the audience gathered in the back will not only form a relatively strong disturbance in appearance but also have a certain movement of their own, which will undoubtedly make things worse.
Therefore, basically all comparison algorithms struggle to handle this scene well, especially the background areas.
By contrast, our method can more fully exploit the roles of different modalities through two symmetrical parallel branches, resulting in satisfactory saliency results.

\subsection{Ablation Study}
In this section, some experiments are conducted to verify the effectiveness of our proposed pipeline and key modules. 

\begin{table}[!t]
\normalsize
\renewcommand\arraystretch{1.1}
	\caption{The ablation verification of CRC and GDR modules on the DAVIS and DAVSOD datasets.}
	\centering
\setlength{\tabcolsep}{0.1cm}{
\begin{tabular}{|c|ccc|ccc|}
				\hline
				\multirow{2}{*}{\textbf{}} & \multicolumn{3}{c|}{\textbf{DAVIS}}   & \multicolumn{3}{c|}{\textbf{DAVSOD}} \\
				\cline{2-7} 
& $F_{\beta}\uparrow$ & $S_{m}\uparrow$ & MAE$\downarrow$ & $F_{\beta}\uparrow$ & $S_{m}\uparrow$ & MAE$\downarrow$ \\ \hline

B &  0.891 & { 0.904} & 0.020  & 0.658 & { 0.746} & 0.082\\
B+GDR &0.895 & 0.907 & 0.019 & 0.663 & 0.750 & 0.080 \\
B+CRC & 0.904 & { 0.913} & 0.017 & 0.671 & { 0.759} &  0.075 \\
B+CRC+GDR  & 0.907 & { 0.919} & 0.016 & 0.678 & { 0.765} & 0.074\\
\hline
			\end{tabular}
		}
	\label{tab:ab}	
\end{table}
\begin{figure}[t]
	\centering
	\includegraphics[width=1\columnwidth]{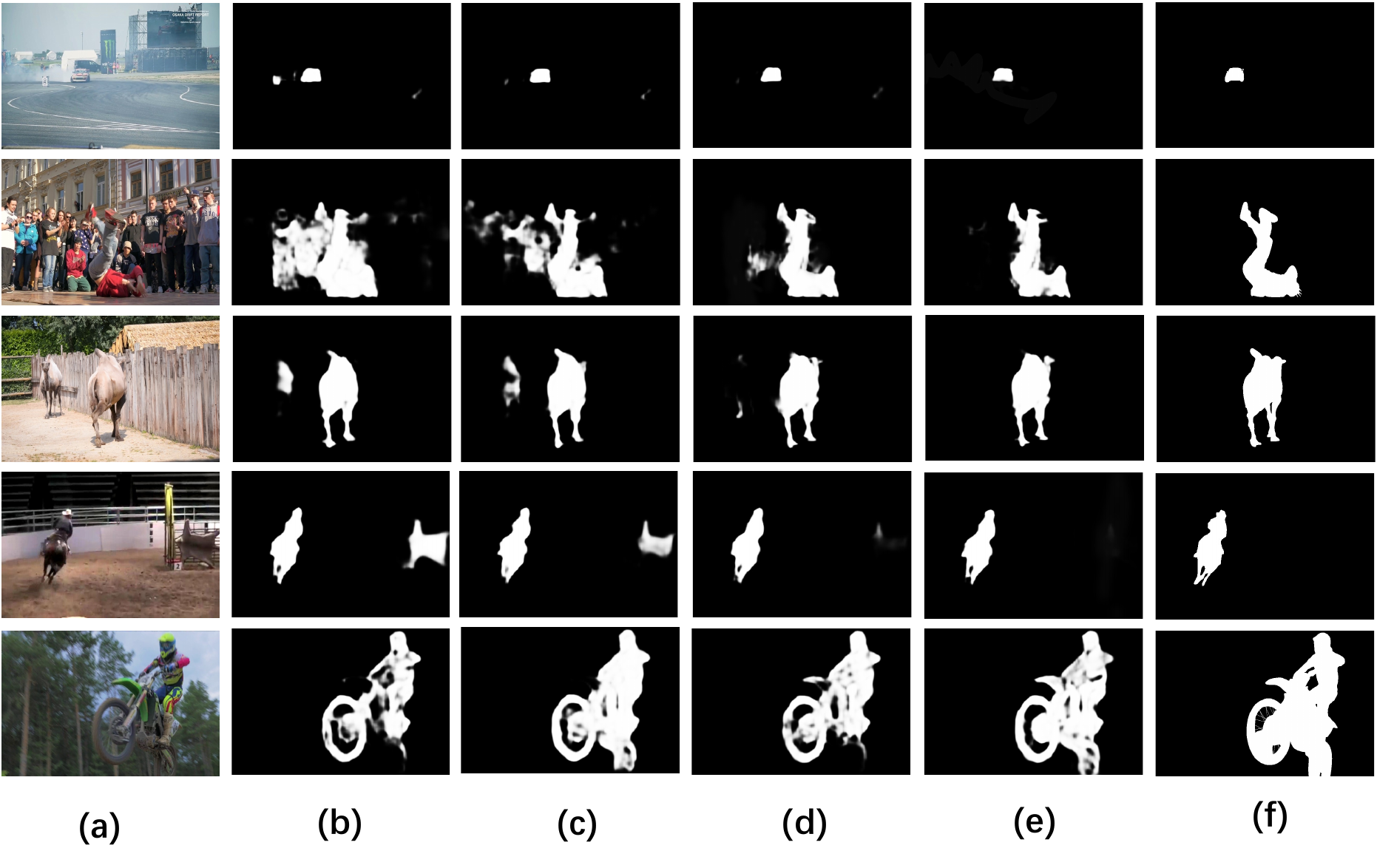}
	\caption {Some visual comparisons of different ablation settings. (a) RGB images; (b) B; (c) B+GDR; (d) B+CRC; (e) B+GDR+CRC; (f) GT.}
	\label{fig9}
\end{figure}
\subsubsection{\textbf{Verification of the GDR and CRC modules}}
We first conduct several experiments to demonstrate the effectiveness of the proposed GDR and CRC modules. Therefore, we keep the IPF module in this part of the experiment.  In order to construct our baseline model, the GDR and CRC modules are removed from the two branches to construct our baseline model. Due to the different importance tendencies of the two modality features in these two branches, we retain the importance sensor in CRC to generate the importance response map. Then, we activate the auxiliary features by multiple them with the importance response map. And the activated auxiliary features are further concatenated with the dominant features in a particular branch. Finally, we accordingly construct a baseline model for our verification (denoted as `B' in Table \ref{tab:ab}). We gradually add CRC and GDR modules into the baseline model for ablation experiments, and the quantitative and qualitative results are shown in Table \ref{tab:ab} and Fig. \ref{fig9}.

Firstly, we add the GDR module to the baseline model (denoted as `B + GDR') to demonstrate the effectiveness of the proposed GDR module. In `B + GDR', the basic setting is similar to ‘B’, but the auxiliary features are activated by the enhanced features from the GDR module. As reported in the second row of Table \ref{tab:ab}, compared with the baseline model on the DAVSOD dataset, the MAE score is improved from 0.082 to 0.080, with a percentage gain of 2.4\%. From Fig. \ref{fig9}(c), after introducing the GDR module, we can see that some background noise can be suppressed slightly, such as the left camel in the third image.
In addition, we also add the CRC module to the baseline model (denoted as `B + CRC') to verify the effectiveness of the CRC module. We can see that introducing the CRC module can boost performance compared with the baseline. Quantitatively, on the DAVIS dataset, introducing the CRC module achieves performance gains of 1.5\% in terms of $F_{\beta}$, and 15.0\% for the MAE score. 
As can be seen in Fig. \ref{fig9}(d), the model with the CRC module has better background suppression ability, such as the items on the right in the fourth image being effectively suppressed. 
Thus, these observations verify that the CRC module can effectively extract useful complementary information from auxiliary modality features and further refine our more important modality features.  Finally, both the GDR and CRC module are introduced into the baseline model to form our full model, which is denoted as `B + CRC + GDR'. Compared with other ablation settings in Table \ref{tab:ab}, our full model achieves the best performance. From Fig. \ref{fig9}(e), we can see that our method achieves a more complete structure, more accurate localization, and clearer background. 

%

\begin{table}[!t]
\normalsize
\renewcommand\arraystretch{1.1}
	\caption{The IPF module verification on DAVIS and DAVSOD datasets. Appearance represents Appearance-Dominated Branch. Motion represents Motion-Dominated Branch.}
	\centering
\setlength{\tabcolsep}{0.125cm}{
\begin{tabular}{|c|ccc|ccc|}
				\hline
				\multirow{2}{*}{\textbf{}} & \multicolumn{3}{c|}{\textbf{DAVIS}}   & \multicolumn{3}{c|}{\textbf{DAVSOD}} \\
				\cline{2-7} 
& $F_{\beta}\uparrow$ & $S_{m}\uparrow$ &  MAE$\downarrow$ & $F_{\beta}\uparrow$ & $S_{m}\uparrow$ &  MAE$\downarrow$ \\ \hline
Parallel-A & 0.905 & { 0.917} & 0.017 & 0.661 & { 0.749} & 0.079\\
Parallel-C & 0.900 & { 0.915} & 0.019 & 0.671 & { 0.761} & 0.076\\
Parallel-F  &  0.901 & { 0.915} & 0.018 & 0.668 &  { 0.757} &  0.076\\
Parallel-IPF  &  0.907 &  { 0.919} & 0.016 & 0.678 &  { 0.765} &  0.074\\
\hline
Appearance  &  0.898 & { 0.911} & 0.018 & 0.662 & { 0.753} & 0.081\\
Motion  &  0.896 &  { 0.910} & 0.019 & 0.648 &  { 0.744} & 0.079\\
\hline
	\end{tabular}
	}
	\label{tab:4}	
\end{table}

 \begin{figure}[t]
 	\centering
 	\includegraphics[width=\columnwidth]{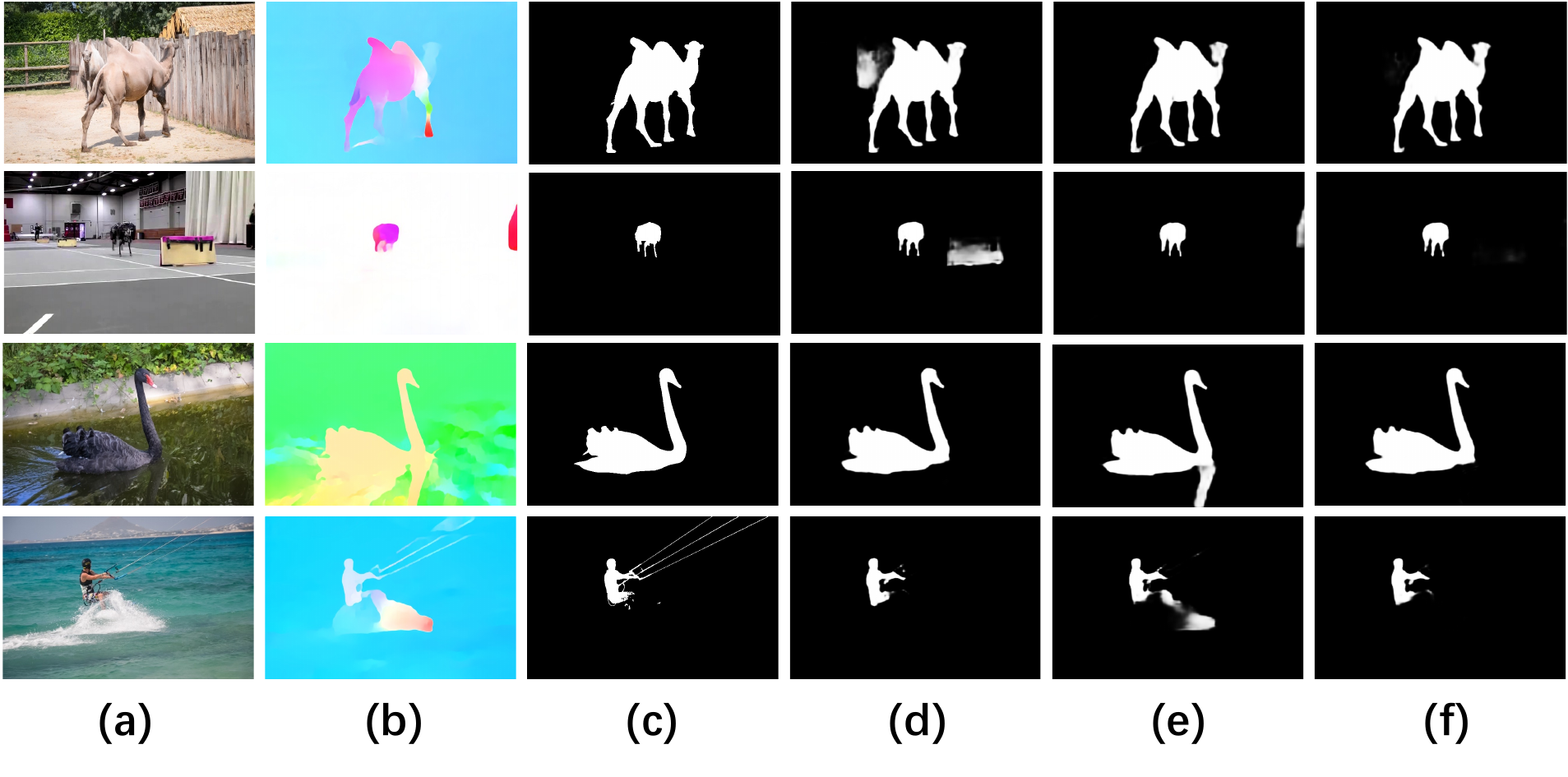}
 	\caption {Some visual comparisons on the output saliency results of Appearance-Dominated Branch, Motion-Dominated Branch, and our IPF. (a) RGB images; (b) Optical flow maps; (c) GT; (d) Saliency results deduced from Appearance-Dominated Branch; (e) Saliency results deduced from Motion-Dominated Branch; (f) Saliency results deduced from our proposed PSNet.
 	}
 	\label{fig8}
 \end{figure}
 
\subsubsection{\textbf{Verification of the IPF module}}
To further verify the effectiveness of our proposed IPF module, we conduct the following ablation models. 
\begin{itemize}
	\item `Parallel-A' denotes that a simple element-wise addition is used to fuse output features from two branches.
\item  `Parallel-C' denotes that concatenation operation is used for fusing two output features.
\item `Parallel-F' denotes that channel-wise attention is used to adaptively fuse the output features of two branches.
\item  `Parallel-IPF' denotes our proposed Importance Perception Fusion module, which uses high-level features in the backbone to sense the importance of two modality data.
\end{itemize}

We retain the multiplication operation for the above experiments to get the common response features between two branch output features and combine the common features with the fused features. As shown in Table \ref{tab:4}, 
our designed IPF module obtains a more robust and generalized final result by considering the different importance of the upper and lower branches in different scenarios. Compared with the `Parallel-F' mode, the percentage gain of the MAE score reaches 11.1\% on the DAVIS dataset and 2.6\% on the DAVSOD dataset, respectively.
In addition, we also report the results for the upper and lower branches (\ie, Appearance-Dominated Branch and Motion-Dominated Branch) under the full-model architecture with the IPF module, as shown in the last two rows of Table \ref{tab:4}. It can be seen that the results of any single branch cannot reach the results with the IPF module, which also illustrates the effectiveness and necessity of our IPF module design. Moreover, in order to further understand the effectiveness of the proposed IPF, some visualization results are shown in Fig. \ref{fig8}. As illustrated in Fig. \ref{fig8}, the motion-dominated branch achieves better saliency results in the top two rows of scenes. While in the last two rows, the appearance-dominated branch achieves better saliency results. And for all these scenes depicted in Fig. \ref{fig8}, our proposed PSNet with the IPF module achieves robust and stable saliency results.

\section{Conclusion}
In this paper, we presented a parallel symmetric network (PSNet) for video salient object detection. Noticing that the importance between appearance cues and motion cues is different under different scenes, we propose to detect saliency via two parallel symmetric branches (\ie, appearance-dominated branch and motion-dominated branch) in an explicitly discriminative way. These two branches have the same structure but regard different modality data as dominant features. Especially, the GDR module is proposed to highlight the multi-scale and multi-layer information, and the CRC module is designed to extract useful information from less important modality data and refine dominant modality data. We also introduce the IPF module to sense the importance weights of two modality data and fuse them adaptively.
Extensive quantitative evaluations and visualization on four benchmark datasets demonstrate that our model achieves promising performance.

\par
\ifCLASSOPTIONcaptionsoff
  \newpage
\fi
{
\bibliographystyle{IEEEtran}
\bibliography{egbib}
}

\end{document}